%% file: samplepaper.tex
\documentclass[runningheads]{llncs}
\usepackage[T1]{fontenc}
\usepackage{cite}
\usepackage{graphicx}
\usepackage{wrapfig}

\usepackage{pifont}
\usepackage{multirow}
\usepackage{booktabs}
\usepackage{amsmath}
\usepackage[table]{xcolor}
\definecolor{headercolor}{RGB}{219,217,226}

\begin{document}
\title{From Moderation to Mediation: Can LLMs Serve as Mediators in Online Flame Wars?}
\titlerunning{From Moderation to Mediation}

\author{Dawei Li\thanks{These authors contributed equally to this work.}\inst{1} \and
Abdullah Alnaibari$^{*}$\inst{1} \and
Arslan Bisharat\inst{2}  \and
Manuel Sandoval\inst{2}  \and\\
Deborah Hall\inst{1}  \and
Yasin Silva\inst{2}  \and
Huan Liu\inst{1}}
\authorrunning{L. Dawei et al.}
%
\institute{Arizona State University, Tempe AZ 85282, USA \and
Loyola University Chicago, Chicago IL 60660, USA \\
\email{daweili5@asu.edu}}
%
\maketitle              
\begin{abstract}
The rapid advancement of large language models (LLMs) has opened new possibilities for AI for good applications. As LLMs increasingly mediate online communication, their potential to foster empathy and constructive dialogue becomes an important frontier for responsible AI research. This work explores whether LLMs can serve not only as moderators that detect harmful content, but as mediators capable of understanding and de-escalating online conflicts. Our framework decomposes mediation into two subtasks: judgment, where an LLM evaluates the fairness and emotional dynamics of a conversation, and steering, where it generates empathetic, de-escalatory messages to guide participants toward resolution. To assess mediation quality, we construct a large Reddit-based dataset and propose a multi-stage evaluation pipeline combining principle-based scoring, user simulation, and human comparison. Experiments show that API-based models outperform open-source counterparts in both reasoning and intervention alignment when doing mediation. Our findings highlight both the promise and limitations of current LLMs as emerging agents for online social mediation.

\keywords{Flaming  \and Large Language Models \and Mediation}
\end{abstract}

\section{Introduction}

\input{Introduction_v1}

\section{Related Work}
\noindent\textbf{AI for Social Good.}
AI for social good is the application of artificial intelligence technologies, such as LLMs, towards solving societal problems. This research domain covers a large number of fields including education, health, environmental sustainability, and economic inequality. For instance, in education, researchers are striving towards precision education, i.e., tailored learning via adaptive digital learning environments. Current efforts employ LLMs for reading and writing assistance \cite{masikisiki_2023}, educational content creation \cite{sami_2022,scarlatos-lan-2023-tree}, and automated feedback and grading \cite{parker2024,morris2024}. LLMs have also been developed to assist in a variety of healthcare-related tasks, such as medical question answering \cite{labrak2024biomistral}, named entity recognition \cite{abishek_2024}, and computer-aided diagnosis \cite{zihao_2024}.

\noindent\textbf{Flame Wars.}
Flame wars are extended exchanges of hostile messages that escalate from disagreement to personal attacks~\cite{moor2010flaming}. The lack of face-to-face cues and online anonymity lower social restraint and intensify hostility~\cite{kiesler1984social, Reicher01011995}. They often unfold through a trigger, reciprocal amplification, and eventual entrenchment. Existing computational work has focused on detecting and filtering toxic content~\cite{steinberger2017flames}, but such moderation may curtail legitimate engagement. Unlike one-sided hate speech, flame wars involve emotionally engaged participants on both sides. Political and identity-related discussions are especially prone to escalation~\cite{koiranen2022political, lee2005flaming}. These dynamics highlight the need for context-aware, constructive intervention—an area our LLM-based mediation framework aims to address.

\section{Task Definition}
\begin{figure}[t]
    \centering
    \includegraphics[width=1\linewidth]{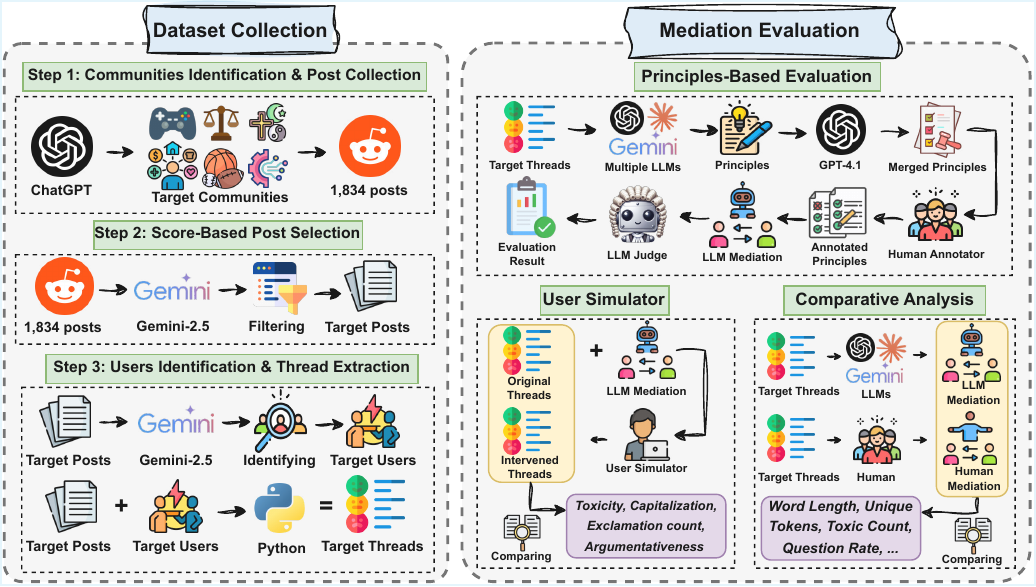}
    \caption{The overview pipeline of our data collection and mediation evaluation process.}
    \label{fig:pipeline}
\end{figure}

Mediation generation aims to move beyond reactive moderation by enabling LLMs to actively understand and intervene in online conflicts. Formally, a discussion thread \( G = \{u_1, u_2, \ldots, u_n\} \) represents a sequence of posts or comments among \( m \) users engaged in disagreement or hostility. The judgment subtask focuses on evaluative reasoning: given \( G \), the model produces an interpretive representation \( J(G) \) that identifies unfair claims, emotional triggers, and points of escalation, while assessing the fairness and relevance of participants’ arguments. The steering subtask focuses on generative mediation: conditioned on \( G \) (and optionally \( J(G) \)), the model generates a mediation output \( S(G) \): a context-aware, empathetic message intended to reduce hostility, acknowledge concerns, and guide participants toward constructive resolution. Together, these subtasks define mediation generation as an interactive process in which an LLM first reasons about conflict dynamics and then produces language that transforms confrontation into cooperation and understanding. Fig.~\ref{fig:pipeline} shows the main processes and tasks of our framework.

\section{Data Collection}

\textbf{Target Communities Identification \& Post Collection.} We chose Reddit,\footnote{https://www.reddit.com/} one of the largest online discussion platforms, as our data source for evaluating if LLMs can serve as mediators, given the open discourse for which the platform is well-known and relative prevalence of flame wars. To identify suitable subreddits, we used ChatGPT~\footnote{https://openai.com/index/chatgpt/} to analyze community dynamics and select forums known for emotionally charged or contentious debates. In total, we collected 1,834 posts, from the past five years, sampled from subreddits representing a wide range of diverse domains, including \textit{Games}, \textit{Lifestyle}, \textit{Religion}, \textit{Social Justice}, \textit{Sports}, and \textit{Technology}—covering subreddits such as \textit{r/gaming}, \textit{r/Parenting}, \textit{r/atheism}, \textit{r/BlackLivesMatter}, \textit{r/football}, and \textit{r/technology}. Descriptive statistic summarizing the sampled subreddits can be found in Table~\ref{tab:dataset_stats}.

\noindent\textbf{Score-Based Post Selection.}
Given the large volume of posts, we designed a filtering process to retain those most relevant to online flame wars. Using Gemini-2.5,~\footnote{https://blog.google/technology/google-deepmind/} each post was automatically scored on a 0--10 scale, where a higher score indicates a greater likelihood that the post contains flame-war interactions in the comments. Only posts receiving scores between 7 and 10 were retained for subsequent analysis, ensuring that the dataset emphasized genuine instances of interpersonal tension and conflict escalation.

\noindent\textbf{Target-user Identification and Thread Extraction.}
We again used Gemini-2.5 to detect the two users with the most flame-war interactions in their comments. After identifying the two users' names, we initially extracted only the direct replies exchanged between them. However, we observed that many relevant subthreads included additional participants whose replies contributed to the escalation or resolution of the conflict. To preserve the full conversational context, we expanded our extraction to include all nested replies within any subtree where both target users appeared, thereby capturing the dynamics of the surrounding dialogue while maintaining a focus on their exchange. After applying this filtering, the final dataset comprised 737 posts with rich conversational structures suitable for mediation generation and principle-based evaluation. Table~\ref{tab:dataset_stats} presents the detailed statistics of our collected dataset.

\begin{table}[h!]
\scriptsize
\setlength{\tabcolsep}{1.5pt}
\centering
\begin{tabular}{lcccccc}
\toprule
\rowcolor{gray!15}
\textbf{Metrics} & \textbf{Games} & \textbf{Lifestyle} & \textbf{Religions} & \textbf{Social Justice} & \textbf{Sports} & \textbf{Technologies} \\
\midrule
\rowcolor{gray!8} \#threads                & 66   & 160  & 155  & 137  & 175  & 44   \\
\rowcolor{white}  \#total comments         & 2696 & 2033 & 3754 & 1980 & 2567 & 2556 \\
\rowcolor{gray!8} \#total users            & 1533 & 844  & 723  & 615  & 1048 & 1337 \\
\rowcolor{white}  \#avg. users per thread  & 24.94 & 7.09 & 7.37 & 7.33 & 7.23 & 33.32 \\
\rowcolor{gray!8} \#avg. comments per thread & 40.85 & 12.71 & 24.22 & 14.45 & 14.67 & 58.09 \\
\bottomrule
\end{tabular}

\caption{Dataset statistics across six sub-datasets. Each sub-dataset corresponds to a topical domain, showing thread count, total comments and users, as well as average users and comments per thread.}
\label{tab:dataset_stats}
\end{table}

\section{Can LLMs Serve as Good Mediators?}

To assess whether large language models can effectively function as mediators in online conflicts, we design a comprehensive evaluation framework consisting of principle-based evaluation, a simulated user environment, and a comparative analysis against human mediation.

\subsection{Principle-based Evaluation}

To ground mediation assessment in interpretable and context-sensitive criteria, we adopt a principle-based evaluation framework aligned with the LLM-as-a-Judge paradigm~\cite{li2025generation}. For each conversation $G = \{u_1, u_2, \dots, u_n\}$, we query three large models—GPT-5, Gemini-2.5, and Claude-4.5—asking each to propose between five and ten principles tailored to that specific discussion, assigning 0–10 scores to each principle. These principles capture fine-grained desiderata for both judgment and steering, which can later guide LLM judges in evaluating mediation quality automatically and efficiently.

\noindent\textbf{Cross-Model Principle Merging.}
The raw principle sets generated by the three models are often overlapping, redundant, or partially misaligned. To consolidate them into a coherent checklist, we use GPT-4.1 as an aggregator. Given the three model-specific lists, GPT-4.1 merges overlapping items, resolves minor wording differences, and removes near-duplicates while preserving important distinctions. This process produces a unified, conversation-specific principle bank $P(G) = \{p_1, p_2, \dots, p_k\}$, which is then human-verified to ensure each principle is faithful to the conversation and practically useful as an evaluation criterion.

\noindent\textbf{Human-Based Principle Verification.}
To ensure quality and consistency, three trained human annotators review each conversation and its merged principle list. They examine the relevance, clarity, and specificity of each principle, performing operations such as \textit{keep}, \textit{edit}, or \textit{delete}, and merging or adding principles when necessary. Each decision is accompanied by a confidence score in $\{1, 2, 3\}$, representing low, moderate, and high confidence. This process yields a high-quality, conversation-specific principle bank suitable for fine-grained mediation assessment.

\noindent\textbf{LLM-as-a-Judge Scoring.}
Finally, we integrate the human-refined principles with model-generated mediation outputs. For each conversation $G$, its principle set $P(G)$ and corresponding mediation output $S(G)$ are jointly provided to an LLM judge $J_{\text{eval}}$, which produces an overall evaluation score: $\text{Score}(G) = J_{\text{eval}}(P(G), S(G)) \in [1, 10]$.
This step enables consistent, interpretable, and scalable evaluation of mediation quality across models.

\subsection{User Simulation Result}
\noindent\textbf{Generation of Intervened Conversations.}
We construct a user simulator that models how conflicting users in online discussions might respond to mediator interventions. Given an original discussion thread $G$, the simulator (here we use Qwen3-4B as the simulator model) generates an intervened thread \( G' \) by inserting a model-generated mediation output \( J(G) \) or \( S(G) \) at an appropriate turn. The simulator then predicts the subsequent user responses conditioned on both \( G \) and \( J(G) \)/\( S(G) \), thereby approximating post-intervention dynamics. This setup enables an evaluation of whether mediation messages can successfully reduce hostility, encourage cooperative tone shifts, or promote topic re-alignment in extended interactions.

\noindent\textbf{Comparison with Original Conversation.}
To quantify the effect of mediation, we compare each intervened thread \( G' \) with its corresponding original version \( G \) without intervention. The comparison focuses on both linguistic and pragmatic changes, including reductions in toxic expressions, improvements in sentiment polarity, and the emergence of collaborative or empathetic language. Automatic metrics such as toxicity reduction, sentiment shift, and engagement balance are computed, supplemented by qualitative inspection of dialogue trajectories. This contrastive setup allows us to measure whether the mediator’s steering effect is constructive and sustainable across user responses.

\subsection{Comparative Analysis}
\noindent\textbf{Human Mediation Collection.}
To establish a reliable reference for LLM-generated mediation quality, we leverage a human mediation dataset from prior work on multilingual moderation and empathetic response generation~\cite{ye2023multilingual}. In the selected dataset, trained human annotators act as mediators in real or sampled conflictual threads drawn from social media. We chose this dataset not only because it contains human-written mediation examples, but also because it includes a Reddit-based subdataset, which aligns closely with the domain of our own collected data.

\noindent\textbf{Metric Design.}
Quantitative analysis spans eleven linguistic and interactional metrics grouped into three categories. 
\textbf{Linguistic complexity} captures how elaborate or readable a reply is, including average words per sentence, average word length, type--token ratio, and Flesch reading ease computed from sentence and syllable counts. \textbf{Interaction dynamics} measure dialogue flow, using question rate (questions per 100 words), engagement balance (questions over questions plus directives), and assertiveness per sentence (directive density normalized by sentence length). \textbf{Interpersonal stance} reflects social orientation and tone, including pronoun bias rate ((you -- we) divided by total words times 100), direct ``you'' references, and toxic word occurrences. 
Additionally, unique token count provides a complementary view of lexical diversity. 
Together, these indicators quantify model--human differences in linguistic form, conversational behavior, and interpersonal affect.

\section{Experiment}

\subsection{Experiment Setting}
\noindent\textbf{Models.}  
We evaluate a total of twelve large language models covering both open-source and API-based systems. The open-source group includes \textit{LLaMA-3.2-3B}, \textit{LLaMA-3.1-8B}, \textit{Qwen2.5-7B}, \textit{Qwen3-1.7B}, \textit{Qwen3-4B}, and \textit{Qwen3-8B}. The API-based group consists of \textit{Claude 3.5-Haiku}, \textit{Claude 4.5-Haiku}, \textit{Claude 4.5-Sonnet}, \textit{GPT-4.1}, \textit{GPT-5}, and \textit{GPT-5.1}. These models represent a balanced selection of architectures and scales, enabling a comprehensive comparison across families and deployment modalities.

\noindent\textbf{Implementation Details.}  
For open-source models, we employ vLLM for efficient local inference and parallelized decoding. All API-based models are accessed through their official provider APIs to ensure consistency and reliability of results. For both the judgment model and the user-simulator model, we use \textit{LLaMA-3-8B} to balance accuracy and computational efficiency. All experiments are conducted on two NVIDIA A100 GPUs (80GB each), ensuring stable runtime and reproducible results across settings.

\subsection{Principle-based Evaluation Result}

\begin{table}[h!]
\small
\setlength{\tabcolsep}{1pt}
\centering
\begin{tabular}{lcccccc ccccccc c}
\toprule
\rowcolor{gray!15}
\textbf{Model} &
\multicolumn{6}{c}{\textbf{Judgment}} &
\multicolumn{6}{c}{\textbf{Steering}} &
\textbf{Avg} \\
\cmidrule(lr){2-7}
\cmidrule(lr){8-13}
 & G & L & R & SJ & S & T 
 & G & L & R & SJ & S & T & \\
\midrule
\rowcolor{gray!8} Qwen2.5-7B & 8.11 & 8.10 & 7.77 & 8.12 & 7.81 & 7.98 & 8.13 & 8.25 & 7.57 & 8.05 & 8.22 & 7.85 & 7.97 \\
\rowcolor{white} Qwen3-1.7B & 8.02 & 8.23 & 7.92 & 8.12 & 7.70 & 7.85 & 8.30 & 8.35 & 7.86 & 8.23 & 8.15 & 7.95 & 8.03 \\
\rowcolor{gray!8} Qwen3-4B & 8.13 & 8.48 & 8.04 & 8.29 & 8.20 & 8.23 & 8.40 & 8.42 & 7.84 & 8.31 & 8.31 & 8.18 & 8.21 \\
\rowcolor{white} Qwen3-8B & 8.28 & 8.38 & 8.01 & 8.32 & 8.21 & 8.28 & 8.36 & 8.45 & 7.88 & 8.33 & 8.35 & 8.30 & 8.23 \\
\rowcolor{gray!8} Llama-3.2-3B & 8.00 & 7.94 & 7.58 & 7.85 & 7.89 & 7.85 & 7.89 & 8.14 & 7.30 & 7.80 & 8.04 & 7.68 & 7.81 \\
\rowcolor{white} Llama-3.1-8B & 7.79 & 7.89 & 7.81 & 7.96 & 7.70 & 7.90 & 8.00 & 8.24 & 7.46 & 7.89 & 8.09 & 7.80 & 7.86 \\
\rowcolor{gray!8} Claude 3.5-Haiku & 8.40 & 8.54 & 8.18 & \textbf{8.55} & 8.25 & \textbf{8.48} & 8.47 & 8.45 & 7.88 & 8.36 & 8.48 & 8.35 & 8.33 \\
\rowcolor{white} Claude 4.5-Haiku & 8.40 & 8.63 & 8.08 & 8.29 & 8.30 & 8.18 & \textbf{8.68} & \textbf{8.76} & \textbf{8.13} & 8.52 & \textbf{8.67} & 8.53 & 8.41 \\
\rowcolor{gray!8} Claude 4.5-Sonnet & 8.23 & 8.49 & 8.06 & 8.49 & \textbf{8.39} & 8.35 & 8.60 & 8.76 & 8.09 & 8.52 & 8.61 & \textbf{8.60} & \textbf{8.41} \\
\rowcolor{white} GPT-4.1 & 8.30 & 8.65 & 8.23 & 8.39 & 8.35 & 8.38 & 8.34 & 8.32 & 7.95 & 8.30 & 8.28 & 8.43 & 8.30 \\
\rowcolor{gray!8} GPT-5 & 8.11 & 8.35 & 7.84 & 8.20 & 8.23 & 7.98 & 8.53 & 8.51 & 7.78 & 8.33 & 8.51 & 8.30 & 8.21 \\
\rowcolor{white} GPT-5.1 & \textbf{8.45} & \textbf{8.51} & \textbf{8.25} & 8.39 & 8.26 & 8.35 & 8.49 & 8.59 & 7.85 & \textbf{8.51} & 8.61 & 8.38 & 8.36 \\
\bottomrule
\end{tabular}

\caption{Principle-based evaluation results across models with bolded highest scores. G: Game, L: Lifestyles, R: Religion, SJ: Social Justice, S: Sports, T: Technologies.}
\label{tab:model_scores}
\end{table}



\noindent\textbf{Models perform differently across topics.} As shown in Table~\ref{tab:model_scores}, models' performance tend to be slightly lower in Religion and Sport, typically ranging from 7.8 to 8.2, while performance in Game and Lifestyle is noticeably stronger, often exceeding 8.4. This pattern suggests that models are more cautious and reserved when dealing with culturally or ethically sensitive subjects like religion and sports. In contrast, topics such as gaming and lifestyle are more neutral and better represented in training data, allowing models to produce smoother and more natural mediation.

\begin{figure*}[h!]
    \centering
    \includegraphics[width=1.0\linewidth]{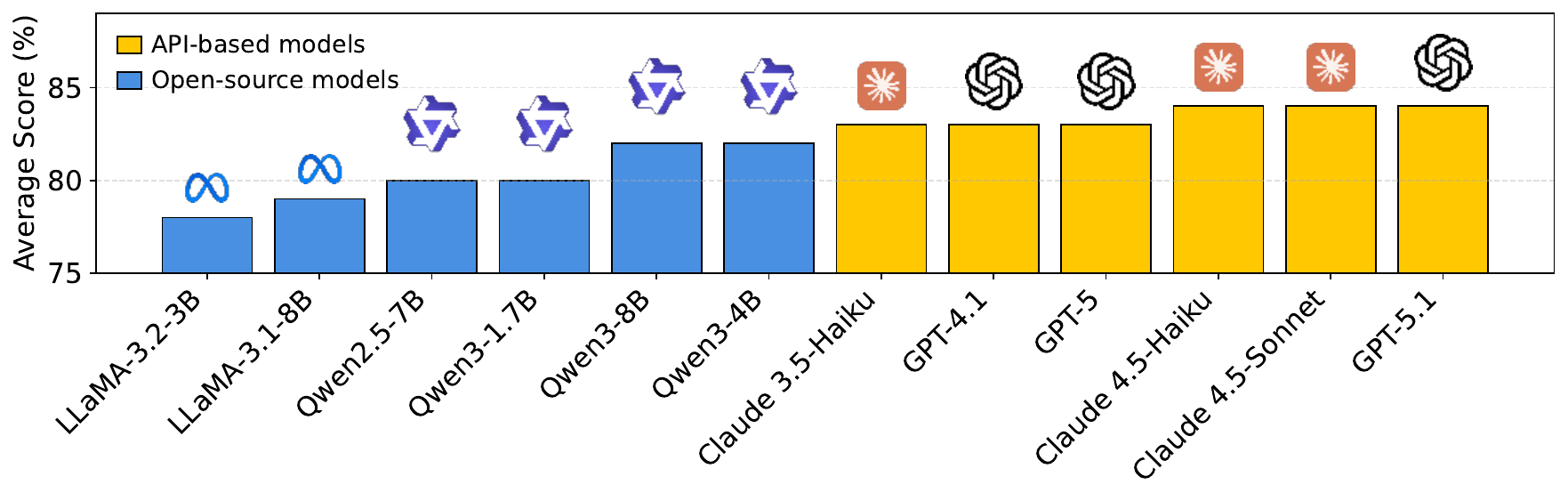}
    \caption{LLM mediation performance leaderboard, averaged on judgment and steering.}
    \label{fig:leaderboard_avg}
\end{figure*}

\noindent\textbf{Closed-source models clearly outperform open-source ones.} In Fig.\ref{fig:leaderboard_avg}, the leaderboard chart, we see that the API-based models like GPT and Claude dominate the top rankings, achieving average scores above 84\%, while open-source models such as Llama and Qwen cluster around the 78–82\% range. This consistent gap suggests that closed-source systems benefit from larger, proprietary datasets and more intensive fine-tuning on alignment and instruction-following. Their commercial training pipelines likely emphasize reliability and user safety, giving them an edge in complex, judgment-oriented evaluations.

\begin{wrapfigure}{r}{0.4\linewidth}
    \centering
    \vspace{-10pt}
    \includegraphics[width=\linewidth]{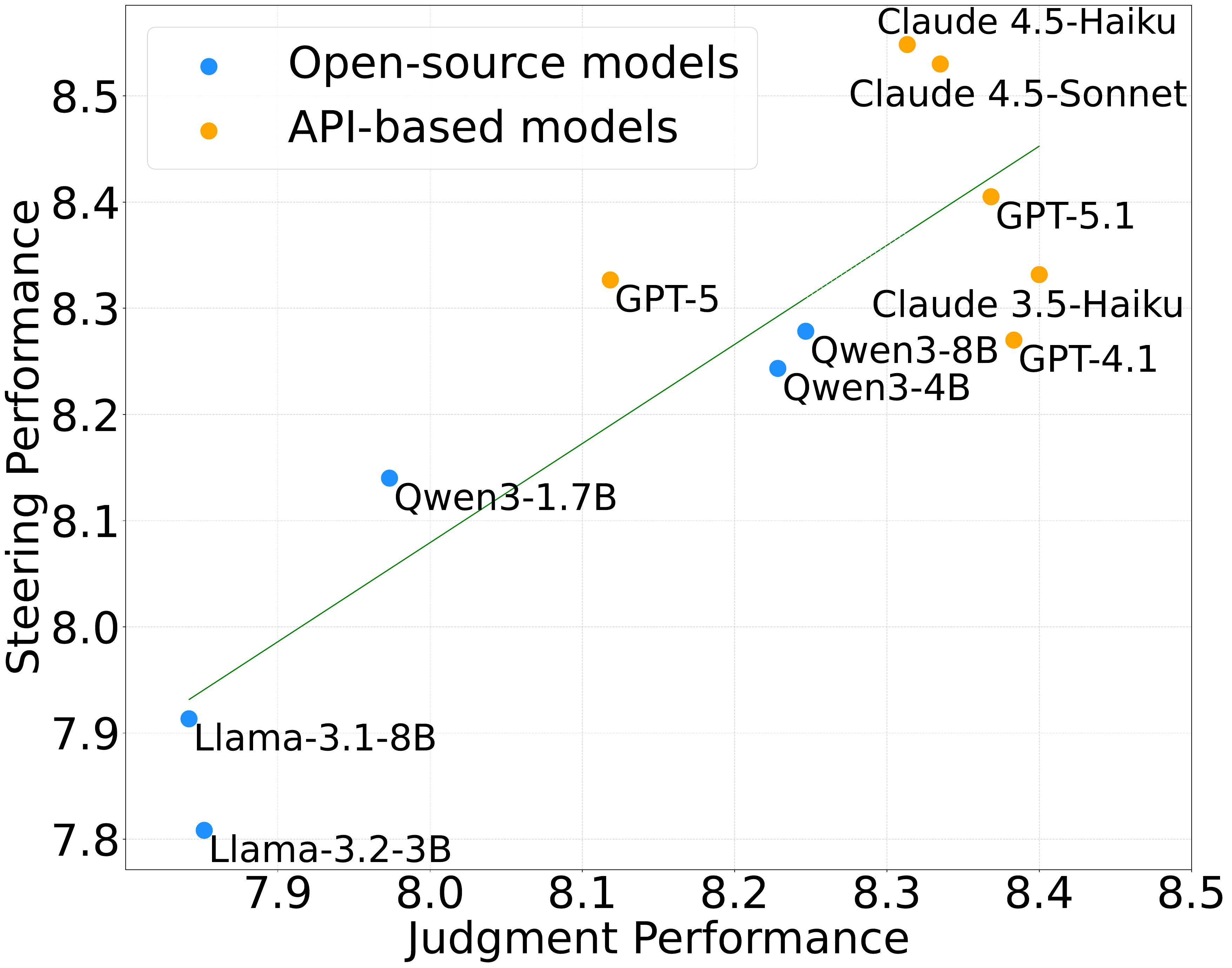}
    \caption{Relationship between judgment and steering scores across models.}
    \label{fig:judgment_vs_steering}
    \vspace{-10pt}
\end{wrapfigure}
\noindent\textbf{Steering and judgment performances are strongly correlated.} Fig.~\ref{fig:judgment_vs_steering}, reveals a clear positive relationship between average steering and judgment scores across all models. Those excelling in judgment also tend to perform well in steering, implying that both tasks draw on shared underlying reasoning and alignment abilities. This consistency suggests that improvements in core mediation skills, like contextual understanding and balanced decision-making, can generalize across multiple evaluation tasks. Future work could explore how to further unify these capabilities, enhancing both interpretability and control in large language models.


\subsection{User Simulation Result}
\begin{table}[h]
\scriptsize
\setlength{\tabcolsep}{4pt}
\centering
\begin{tabular}{lccccc}
\toprule
\rowcolor{gray!15}
\textbf{Model} & \textbf{Toxicity} & \textbf{Capitalization} & \textbf{Exclamation} & \textbf{Argumentativeness} & \textbf{Avg} \\
\midrule
\rowcolor{gray!8} Qwen2.5-7B    & \textbf{39.92} & \textbf{0.30} & 17.31 & \textbf{6.42} & 16.09 \\
\rowcolor{white}  Qwen3-1.7B    & 30.27          & \textbf{0.30} & 19.65 & 6.01          & 14.06 \\
\rowcolor{gray!8} Qwen3-4B      & 25.42          & \textbf{0.30} & 19.19 & 6.35          & 12.81 \\
\rowcolor{white}  Qwen3-8B      & 26.27          & \textbf{0.30} & 19.19 & 6.08          & 13.00 \\
\rowcolor{gray!8} Llama-3.1-8B  & 36.85          & \textbf{0.30} & \textbf{20.42} & 6.18     & 15.94 \\
\rowcolor{white}  Llama-3.2-3B  & 38.50          & \textbf{0.30} & 20.00 & 5.99          & \textbf{16.10} \\
\bottomrule
\end{tabular}
\caption{User simulation results on open-source LLMs. Toxicity measures the rate of negative lexicon; Capitalization reflects excessive uppercase ratio; Exclamation counts exclamation marks; Argumentativeness measures disagreement vs. politeness balance.}
\label{tab:user_simulation}
\end{table}

The results in Table~\ref{tab:user_simulation} show that mediation has varying effectiveness across different linguistic aspects. The reduction in toxicity and exclamation usage suggests that the mediation process successfully promotes more neutral and composed responses. This can improve conversational tone and make model interactions feel more balanced and less emotionally charged. However, the argumentativeness scores change only slightly, indicating that mediation has a weaker influence on discourse-level behaviors such as disagreement management.

Interestingly, the trends differ between the user simulation setting and the main experiment, suggesting that models behave differently when generating both sides of an interaction versus responding to real user input. This highlights the importance of evaluating mediation not only as a process (how mediation is applied) but also in terms of its outcomes (the actual communicative effects in realistic contexts).

\subsection{Comparative Analysis Result}
\begin{figure*}[h!]
    \centering
    \includegraphics[width=1.0\linewidth]{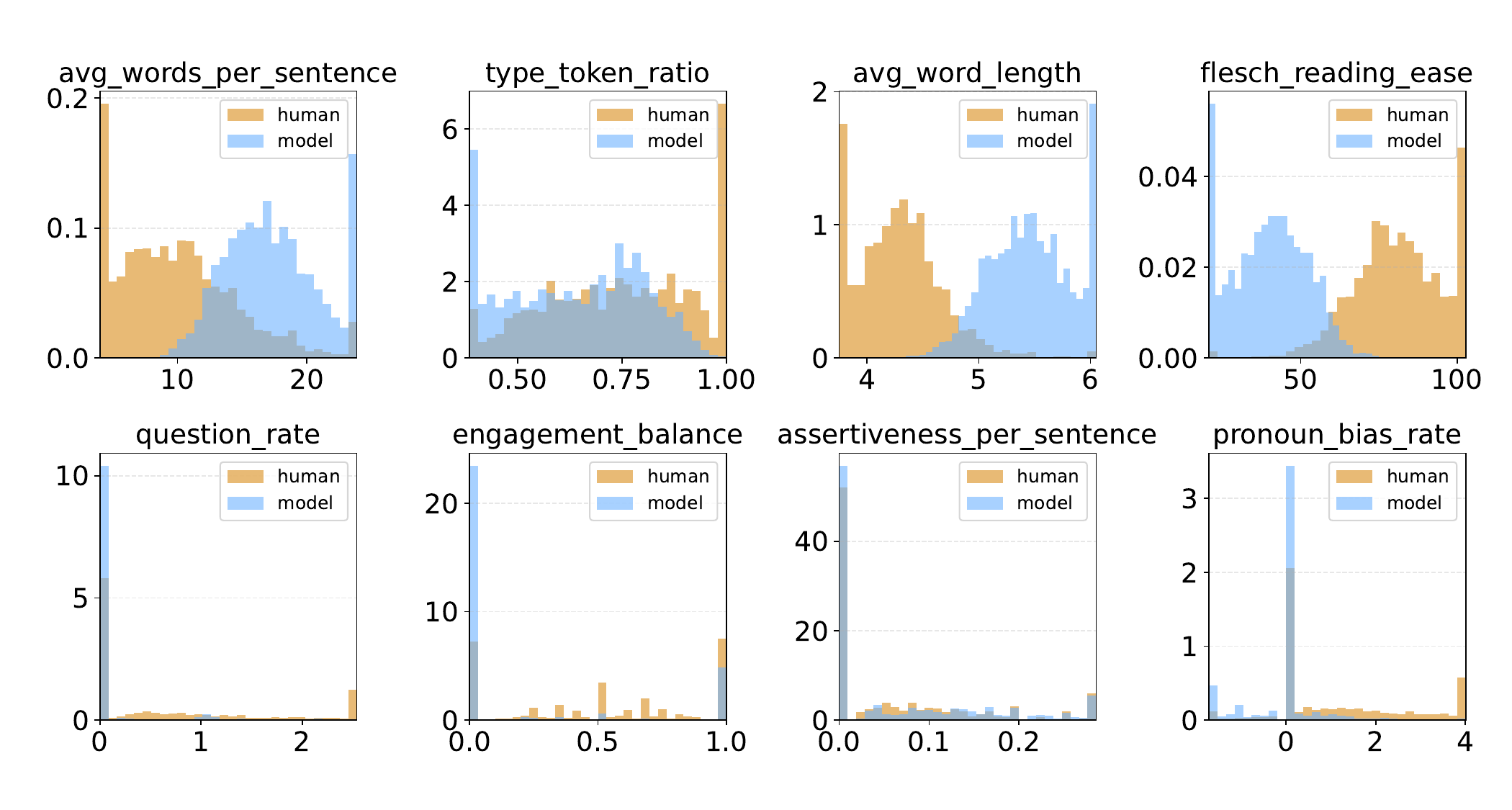}
    \caption{Distributions of linguistic and interactional metrics.}
    \label{fig:feature_distribution}
\end{figure*}
\begin{wrapfigure}{r}{0.6\textwidth}
    \centering
    \includegraphics[width=\linewidth]{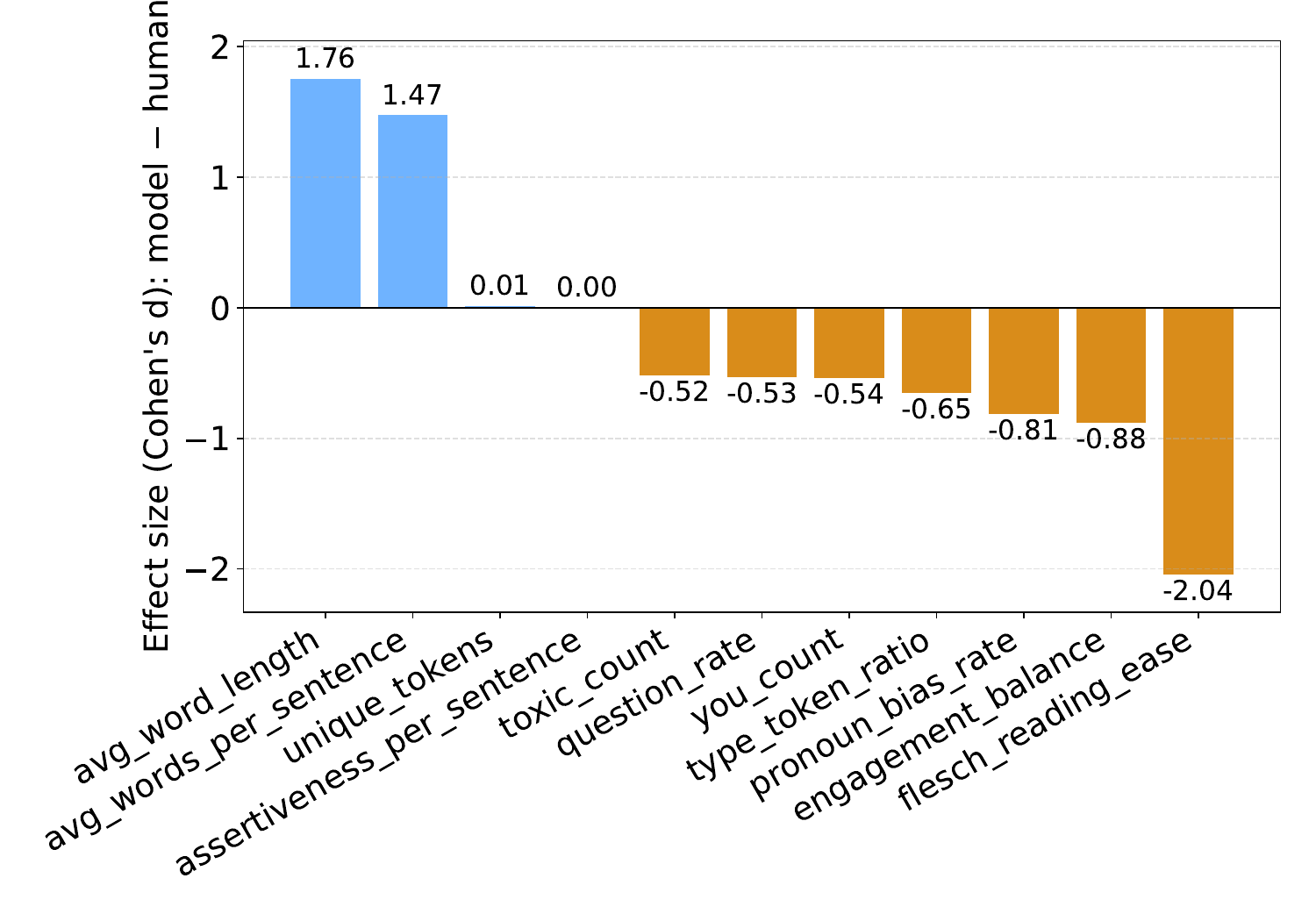}
    \caption{Effect sizes of model–human differences.}
    \label{fig:size_comparison}
\vspace{-10pt}
\end{wrapfigure}
Fig.~\ref{fig:feature_distribution} demonstrates the distribution difference between human- and LLM-generated mediation. Across lexical, structural, readability, and interactional dimensions, human- and LLM-generated mediations exhibit distinct yet complementary linguistic profiles. LLM outputs tend to be substantially longer and lexically denser (d $\approx$ 1.5–1.7), reflecting a more formal and elaborated register. However, this expansion corresponds with decreased readability (Flesch d $\approx$ –2.0), indicating that human mediations remain considerably more accessible to readers. Human texts further display higher lexical diversity and stronger dialogic engagement, evidenced by greater question frequency and more balanced interactional tone.

We also demonstrate the effect sizes of model–human differences. As Fig.~\ref{fig:size_comparison} shows, pronoun usage underscores differences in stance and interpersonal orientation. Human mediators show higher “you” frequency, adopting a direct and addressive style, whereas LLMs favor more neutral or collective pronoun patterns. This contributes to lower toxicity proxies in LLM outputs, suggesting reduced confrontational or accusatory tendencies. Assertiveness per sentence remains near parity, implying that both maintain comparable directive moderation.


\section{Conclusion}
This work explored whether large language models can act not just as moderators but as mediators capable of understanding and resolving online conflicts. Through a combination of principle-based evaluation, user simulation, and comparative analysis, our findings show that current LLMs demonstrate promising yet uneven mediation abilities. Closed-source models like GPT and Claude consistently outperform open-source systems in both judgment and steering tasks, suggesting that fine-tuned alignment and large-scale instruction optimization play a key role in effective mediation. Moreover, the strong correlation between steering and judgment performance highlights that both tasks rely on shared reasoning and empathy-related capabilities. This implies that improvements in one area can benefit the other, reinforcing the idea of a unified mediation competence within LLMs.

\section*{Acknowledgement}
This work was supported by National Science Foundation Awards No. 2435164 and 2435165, and a Google Award for Inclusion Research. The views and conclusions contained in this document are those of the authors and should not be interpreted as representing official policies, either expressed or implied, of NSF.
%
%
%
\bibliographystyle{splncs04}
\bibliography{mybibliography}

\end{document}

%% file: Introduction_v1.tex
The rapid advancement of large language models (LLMs) has reshaped how humans communicate, learn, and collaborate. Beyond traditional natural language tasks such as translation or summarization~\cite{gao2023human}, LLMs are increasingly being applied to domains that serve social good, supporting applications such as education~\cite{sami_2022,scarlatos-lan-2023-tree}, healthcare~\cite{labrak2024biomistral,abishek_2024} and mental well-being~\cite{Hua2025ASR}. As these systems become embedded in online spaces where people interact and debate, a new frontier emerges: using LLMs not just as tools for automation but as agents that foster empathy, fairness, and constructive dialogue in digital communities~\cite{xu2025multiagentesc}.

One pressing challenge in this landscape is the persistence of online hostility, often manifested through flame wars: heated, multi-turn exchanges of personal attacks and emotional escalation~\cite{moor2010flaming}. The lack of face-to-face cues and the anonymity of the medium lower social restraint and amplify anger or frustration, while group identity dynamics can further polarize discussions~\cite{kiesler1984social}. While computational methods have made progress in detecting toxicity or flagging abusive content~\cite{steinberger2017flames}, they remain largely reactive, focused on moderation through removal or punishment rather than prevention or repair. As a result, online platforms can suppress harmful speech without addressing the underlying social dynamics that cause it, leaving space for hostility to resurface.

Our work aims to explore a different approach to intervention: mediation rather than moderation. Instead of judging messages as harmful, we investigate whether an LLM can track a conflict and help de-escalate it. We view LLMs as potential mediators that can interpret multi-turn conversations, assess the fairness and relevance of the arguments being made, and generate responses that acknowledge concerns, reduce hostility, and guide the exchange toward a more constructive path. Achieving this requires reasoning about context, emotion, escalation patterns, and conversational norms \cite{madani2025esc}. To achieve this goal, we divide the task into two complementary components. Judgment involves evaluating the interaction to identify unfair claims, emotional triggers, and points of escalation. Steering involves generating a message that calms the exchange and helps participants focus on the source of their disagreement rather than attacking each other.

To examine whether LLMs can indeed serve as capable mediators, we construct a large-scale dataset of real-world flame wars from Reddit and develop a multi-stage evaluation pipeline. Our framework integrates three perspectives: \ding{182} principle-based evaluation, grounding mediation quality in fine-grained conversational principles; \ding{183} user simulation, modeling post-intervention behavior to assess whether LLM-generated mediations reduce hostility; and \ding{184} human comparative assessment, benchmarking model interventions against human-written mediations. From extensive experiments, we found that API-based models consistently outperform open-source systems in both judgment and steering alignment, exhibiting greater coherence and stability across categories. Furthermore, mediation analysis reveals that LLMs effectively reduce toxicity and emotional intensity in simulated interactions, while comparative evaluation highlights their strengths in neutrality and structural precision but relative weakness in readability and human-like engagement.